\documentclass[sigconf]{acmart}

\AtBeginDocument{%
  }

\setcopyright{acmcopyright}
\copyrightyear{2023}
\acmYear{2023}
\acmDOI{}

\acmConference[ICEA 2023]{2023 International Conference on Intelligent Computing and Its Emerging Application}{Dec 13--14,
  2023}{Yanchao, Kaohsiung}

\acmPrice{}
\acmISBN{}

\usepackage{bm}
\usepackage{algorithm,algpseudocode}
\usepackage{graphicx}
\graphicspath{{figs/}}
\usepackage{subcaption}
\usepackage{multirow}
\usepackage{booktabs}
\usepackage{stfloats}
\usepackage[justification=justified]{caption}
\begin{document}

\title{Roadside LiDAR Assisted Cooperative Localization for Connected Autonomous Vehicles}

\author{Yuze Jiang}
\affiliation{%
  \institution{The University of Tokyo}
  \state{Hongo, Bunkyo-ku}
  \city{Tokyo}
  \country{Japan}
}
\email{uiryuu@g.ecc.u-tokyo.ac.jp}

\author{Ehsan Javanmardi}
\affiliation{%
  \institution{The University of Tokyo}
  \state{Hongo, Bunkyo-ku}
  \city{Tokyo}
  \country{Japan}
}
\email{ejavanmardi@g.ecc.u-tokyo.ac.jp}

\author{Jin Nakazato}
\affiliation{%
  \institution{The University of Tokyo}
  \state{Hongo, Bunkyo-ku}
  \city{Tokyo}
  \country{Japan}
}
\email{jin-nakazato@g.ecc.u-tokyo.ac.jp}

\author{Manabu Tsukada}
\affiliation{%
  \institution{The University of Tokyo}
  \state{Hongo, Bunkyo-ku}
  \city{Tokyo}
  \country{Japan}
}
\email{mtsukada@g.ecc.u-tokyo.ac.jp}

\author{Hiroshi Esaki}
\affiliation{%
  \institution{The University of Tokyo}
  \state{Hongo, Bunkyo-ku}
  \city{Tokyo}
  \country{Japan}
}
\email{hiroshi@wide.ad.jp}


\begin{abstract}
Advancements in LiDAR technology have led to more cost-effective production while simultaneously improving precision and resolution. As a result, LiDAR has become integral to vehicle localization, achieving centimeter-level accuracy through techniques like Normal Distributions Transform (NDT) and other advanced 3D registration algorithms. Nonetheless, these approaches are reliant on high-definition 3D point cloud maps, the creation of which involves significant expenditure. When such maps are unavailable or lack sufficient features for 3D registration algorithms, localization accuracy diminishes, posing a risk to road safety. To address this, we proposed to use LiDAR-equipped roadside unit and Vehicle-to-Infrastructure (V2I) communication to accurately estimate the connected autonomous vehicle's position and help the vehicle when its self-localization is not accurate enough. Our simulation results indicate that this method outperforms traditional NDT scan matching-based approaches in terms of localization accuracy.
\end{abstract}




\maketitle

\section{Introduction}
Vehicle localization is vital for effective navigation, especially in intricate urban settings. Although the majority of modern navigation systems rely on Global Positioning System (GPS), its inherent limitations have spurred the development of advanced GPS methodologies, such as Differential GPS, and Real Time Kinematic GPS, to enhance accuracy. When augmented with these techniques, GPS-based systems can achieve meter-level precision \cite{Kuutti2018-rs}. However, in the context of autonomous driving, pinpoint localization is necessary, requiring errors less than 30cm to correctly identify lanes \cite{Levinson2007-gm}. Recent advancements in 3D point cloud HD map-based localization methods have surpassed GPS in accuracy, making autonomous driving increasingly viable. When vehicles equipped with LiDAR (Light Detection and Ranging) sensors drive through places where such HD maps are available, registration algorithms like Normal Distributions Transform (NDT) \cite{Biber2003-xx} or Iterative Closest Point (ICP) \cite{Arun1987-zt} are utilized for vehicle localization. These approaches can achieve centimeter-level localization accuracy \cite{Kuutti2018-rs} in favorable situations in terms of identifiable features in the map. However, the absence or repetition of such map features compromise localization performance \cite{Javanmardi2018-us}.

With the development of Vehicle Ad-hoc Network (VANET), vehicles can communicate with other connected vehicles and connected infrastructure \cite{Mansour2018-tz}. To further enhance the localization accuracy and reliability, numerous studies have explored cooperative localization strategies that utilize Vehicle-to-Vehicle (V2V) and Vehicle-to-Infrastructure (V2I) communications \cite{Yang2020-wz} \cite{Ou2019-xv} \cite{Khattab2015-cy}. The stationary nature of infrastructure serves as a major advantage for localization, which make vehicles easily determine their absolute position based on their relative position to the infrastructure. In this paper, we aim to improve the localization of connected autonomous vehicles by leveraging V2I communication and roadside LiDAR. We proposed a novel method to communicate the 2D bounding box size of the connected vehicle to the LiDAR-equipped RSU, which significantly improves the accuracy for the roadside LiDAR to estimate the location of the connected vehicle. We evaluated the effectiveness by comparing the result with NDT scan matching-based localization in a realistically simulated environment.


The key contributions of this paper are outlined below:
\begin{itemize}
    \item Our study explored the potential of utilizing roadside LiDAR for object detection to provide localization for connected autonomous vehicles.
    \item We introduced a method to enhance the precision of estimating the central point of a vehicle from roadside infrastructure, utilizing V2I communications.
    \item Through simulation experiments, we demonstrated that our proposed approach surpasses the conventional NDT-based self-localization for vehicles in certain scenarios.
\end{itemize}

The remainder of this paper is organized as follows: Section~\ref{related} reviews existing literature on cooperative localization strategies, with a focus on techniques that incorporate roadside units and LiDAR. Section~\ref{method} outlines the experimental design and the proposed method. In Section~\ref{experiment}, we detailed our realistic experiment and analysis its result. Concluding remarks and future research directions are presented in Section~\ref{conclusion}.

\section{Related Work}\label{related}
\subsection{LiDAR-Based Vehicle Self-Localization}
Though NDT and ICP are both widely used algorithms for point cloud registration and vehicle localization, NDT is generally superior to ICT in terms of accuracy and robustness~\cite{Pang2018-aj}. In a study by Kan \textit{et al.}~\cite{Kan2021-xy}, the efficacy of map-based NDT scan matching was assessed under conditions where LiDAR scans were partially occluded. The results revealed that localization remained stable until the occlusion rate exceeded 25\%, and only faltered when the rate surpassed 80\%, underscoring the resilience of NDT scan matching. Despite its strengths, the success of NDT-based localization is contingent upon the map's granularity~\cite{Pang2018-aj} and the availability of distinctive features~\cite{Javanmardi2018-us}. Consequently, there exist places where we cannot merely depend on map-based localization methods due to lack of features or features being repeated over time.

\subsection{Cooperative Localization Methods}
For self-localization, vehicles usually only use data from the onboard sensors. Cooperative localization, however, expands this scope by incorporating external information, communicated through the VANET between connected agents. Cooperative localization methods typically falls into two main types: V2V-based and V2I-based.

V2V-based cooperative localization methods involve multilateration, using inter-vehicle distance data and individual vehicle self-localization outcomes to determine a subject vehicle's relative position. In \cite{Fujii2011-dm}, Fujii \textit{et al.} proposed a multiple-vehicle system in which the vehicles measure and share the distances to other vehicles together with their current position. Upon receiving the shared information, a vehicle can improve its localization accuracy by over 50~\% compared to self-localization. Yang \textit{et al.} developed a Multi-Sensor Multi-Vehicle framework \cite{Yang2020-wz} which enables data from advanced sensors, including LiDAR, to be fused via V2V communication. Nevertheless, the efficiency of V2V cooperative localization is fundamentally limited by each vehicle's self-localization accuracy. The effectiveness of the whole system deteriorates when all vehicles cannot perform self-localization accurately enough, for example, due to environmental factors.

On the other hand, stationary RSUs can be used as reference anchors in the V2I-based cooperative localization systems, significantly narrowing the source of error. Existing literature mainly employs radio positioning techniques, which not only demand precise time synchronization between the RSUs and the vehicles but are also susceptible to multipath effects~\cite {Nguyen2023-kt}. Ou \textit{et al.} proposed a system employing RSUs with directional antennas \cite{Ou2019-xv}. Once two or more RSUs detect a vehicle, it can refine its position through geometric calculations. However, this approach struggles with complex maneuvers such as sharp turns and U-turns. Liu \textit{et al.} developed a collaborative positioning algorithm that merges received signal strength indication (RSSI) from RSUs, GPS data, dead reckoning, and V2I communication \cite{Liu2022-ip}. While the algorithm outperforms GPS-only methods, its meter-level accuracy under realistic scenarios falls short for tasks like lane identification, rendering it inadequate for autonomous driving.

\subsection{Vehicle Detection and Tracking via Roadside Infrastructure}
Infrastructure plays a vital role in the VANET. As stipulated by the ESTI (European Telecommunications Standards Institute) standard for Collective Perception Service (CPS) \cite{Etsi2023-ao}, infrastructure has the role of obtaining information from mounted stationary sensors and broadcasting the information via Collective Perception Message (CPM) to other connected agents nearby. Therefore, the infrastructure is responsible for tracking and monitoring the objects (including vehicles and pedestrians) on the road. 

Zhao \textit{et al.} \cite{Zhao2019-jq} conducted a real-world experiment where roadside LiDAR is used to detect, track the trajectories, and estimate the speed of the vehicles and pedestrians at an intersection. In this study, the authors evaluated the accuracy of the estimated speed of the vehicles by comparing it with the speed measured from the onboard devices. Subsequently, Zhang \textit{et al.} \cite{Zhang2020-fs} improved the roadside LiDAR-based speed tracking performance by introducing centroid-based tracking. Despite their contributions, neither study intended to leverage the roadside LiDAR for enhancing the localization accuracy for the connected vehicles.

\begin{figure}[h]
    \centering
    \includegraphics[width=0.8\linewidth]{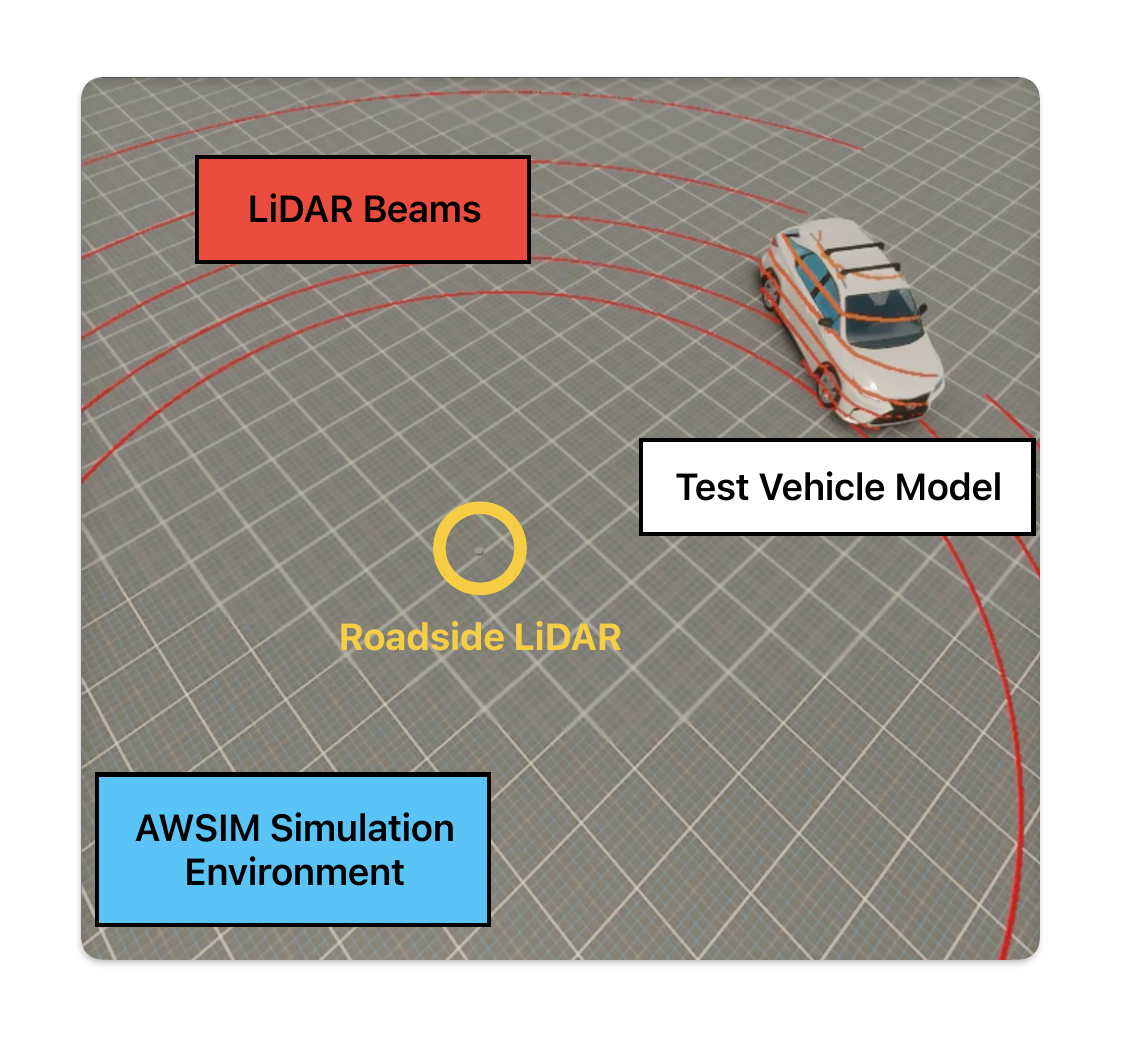}
    \caption{Setup of the Pilot Experiment to Generate Heat Map}
    \label{fig:heatmap_setup}
\end{figure}

\section{Methodology}\label{method}

Despite significant advancements in LiDAR technology enabling precise vehicle self-localization, its dependency on detailed 3D point cloud maps with identifiable features presents limitations in reliability under certain conditions. The challenge posed by maps with insufficient features is not easily resolved. However, the deployment of LiDAR-equipped roadside infrastructure emerges as a viable solution to these limitations. Our proposal, therefore, focuses on investigating the role of this infrastructure in enhancing localization accuracy.

\subsection{Feasibility Study}\label{init}
In the initial phase of our research, we carried out a preliminary study to investigate the capabilities of roadside LiDAR for vehicle localization in optimal conditions. Heat maps of localization error were generated to visualize the results. We employed AWSIM~\cite{TIER_IV_inc_undated-to}, a Unity-based digital twin simulator tailored for autonomous driving research, as our primary simulation environment. AWSIM uses RobotecGPULidar (RGL) \cite{Ai_undated-ub} as its LiDAR simulator, which is an efficient CUDA-powered LiDAR simulator that can simulate the LiDAR behavior quickly and accurately. The host machine uses an AMD Ryzen 9 5950X CPU and NVIDIA RTX 3090 GPU.

\begin{figure*}
    \centering
    \begin{subfigure}[b]{0.475\textwidth}
        \centering
        \includegraphics[width=\textwidth]{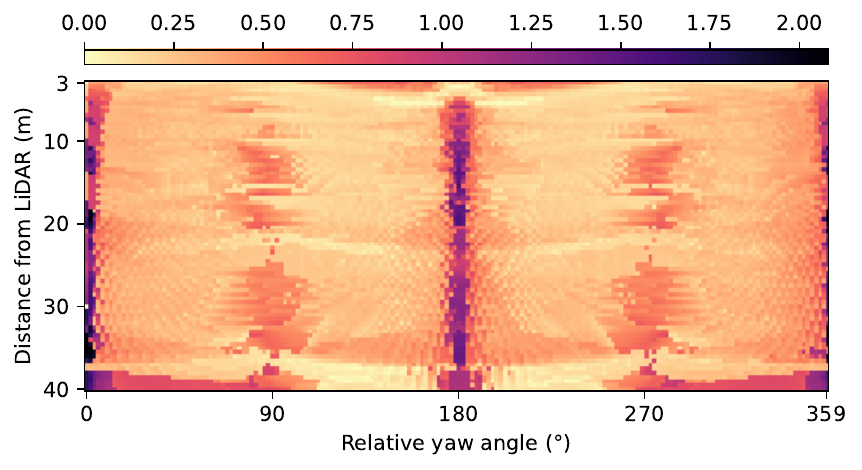}
        \caption{Error of the Center Point in Distance (m)}
        \label{fig:error_dist_original}
    \end{subfigure}
    \hfill
    \begin{subfigure}[b]{0.475\textwidth}  
        \centering 
        \includegraphics[width=\textwidth]{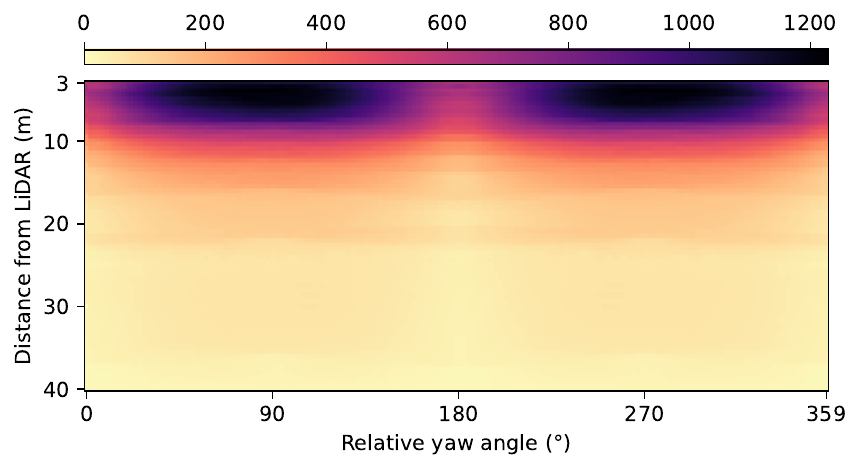}
        \caption{Number of Points}
        \label{fig:num_of_points}
    \end{subfigure}
    \\[2ex]
    \begin{subfigure}[b]{0.475\textwidth}   
        \centering 
        \includegraphics[width=\textwidth]{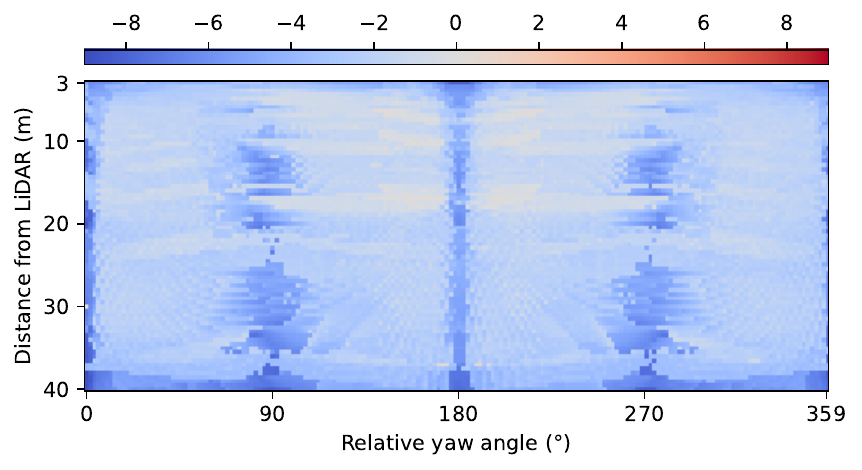}
        \caption{Error in the Bounding Box Size ($m^2$)}
        \label{fig:error_in_bb}
    \end{subfigure}
    \hfill
    \begin{subfigure}[b]{0.475\textwidth}   
        \centering
        \includegraphics[width=\textwidth]{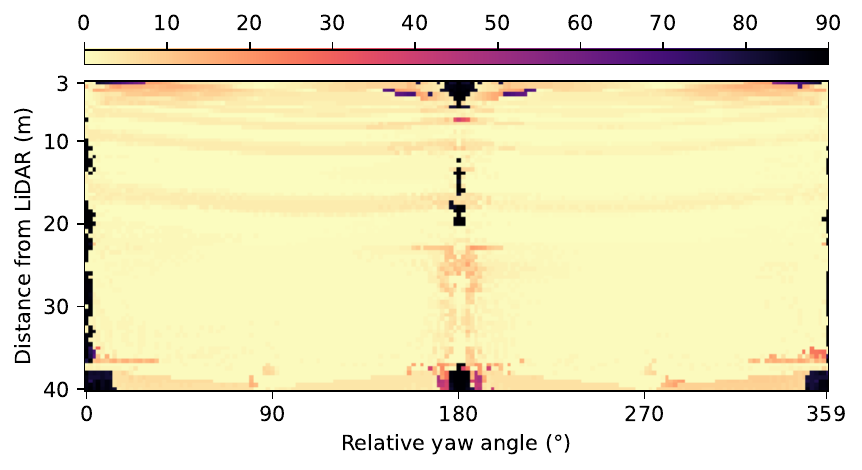}
        \caption{Error in Yaw Angle (°)}
        \label{fig:error_in_angle}
    \end{subfigure}
    \caption{Error and number of points when using the original L-shape fitting algorithm on VLP-16}
    \label{fig:error_figs_original_algo}
\end{figure*}

We set up the experiment in an empty scene in the AWSIM Unity project, as illustrated in Figure~\ref{fig:heatmap_setup}, a LiDAR sensor was positioned at a static point in the air to function as a RSU. The LiDAR perceives the surrounding environment and publishes to ROS \cite{Macenski2022-xv} topics. A test vehicle rotates around its center point at a constant rate. Upon completing a full 360° rotation, the vehicle then moves 0.5 m away from the RSU. This procedure allowed us to collect point clouds for various combinations of distances from the vehicle to RSU and the vehicle's yaw angles. We then employed these point clouds to estimate the vehicle's position, contrasting this with the ground truth to assess the localization error for each combination of distance and angle. The detailed configuration is shown in Table~\ref{tab:heatmap_setup}.

\begin{table}
\centering
\caption{Pilot Experiment Parameters}
\begin{tabular}{l|l}
\hline
\hline
Test Vehicle Model        & Lexus RX450h 2015 \\
Vehicle Rotation Speed    & 20° per second    \\
LiDAR Model               & Velodyne VLP-16   \\
LiDAR Height              & 2 m                \\
LiDAR Reporting Frequency & 10 Hz              \\
\hline
\hline
\end{tabular}
\label{tab:heatmap_setup}
\end{table}

We filtered out zero-height points as background filtering from the collected point cloud data, leaving only vehicle points. Then, the L-shape fitting algorithm is used to estimate 2D bounding boxes of the vehicle points. L-shape Fitting Algorithm was first proposed by Zhang \textit{et al.}~\cite{Zhang2017-be} and is currently used as the default bounding box estimation algorithm in Autoware~\cite{TIER_VI_inc_undated-ur}. The algorithm evaluates various placement angles and their closeness scores to find the best-fitting rectangle. The center point of the bounding box is used as the vehicle's estimated location and compared with the ground truth to generate the error heat map. The yaw angle was estimated based on the orientation of the rectangle's longer edge.

Data was gathered for distances ranging from 3 m to 40 m between the vehicle and RSU. Starting the experiment from 0 m was meaningless due to the VLP-16 LiDAR's 30° vertical FoV~\cite{Roriz2022-gy}, which creates a detection gap directly underneath it. Post L-shape fitting, we assessed various errors: distance error for the bounding box center, bounding box size error, the number of points caused by the vehicle perceived by the roadside LiDAR, and yaw angle error at each given distance and angle, as detailed in Figure~\ref{fig:error_figs_original_algo}. The following observations were made from the figures.

First, the localization accuracy drops sharply when the vehicle is relatively parallel and perpendicular to the LiDAR, however, the accuracy is not proportional to the number of points. In Figure~\ref{fig:error_in_angle}, there are some black spots, which means the angle estimation is off by 90°. This is because the LiDAR beams cannot penetrate the vehicle. L-shape fitting works well when two edges of the vehicle are sensed. However, since the LiDAR beams cannot penetrate the vehicle, we can only get a partial point cloud of the vehicle at a time.  When the vehicle's yaw angle is 0° (or 360°), meaning the vehicle's front or back is directly pointing to the LiDAR, only half of the vehicle is sensed. Since we solely estimate the yaw angle by the length of the edges of the bounding box, if the shorter edge in the estimated bounding box is the longer edge of the vehicle, we can observe the off-by-90° error.

Besides, the pattern of localization inaccuracy closely aligns with the error pattern for bounding box size as seen in Figure~\ref{fig:error_in_bb}. This suggests that errors in bounding box dimensions correlate with errors in the estimated position of the vehicle's center. Considering that the size of the vehicle does not change over time, we proposed to leverage V2I communication and let the connected vehicle tell the RSU about its dimension information so that the RSU can compute a more accurate localization result to help the vehicle self-localization.

\subsection{Position Refinement by Bounding Box Size Correction}

\begin{algorithm}[t]
\caption{Bounding Box Size Correction Algorithm}\label{alg:size_correction}
\begin{algorithmic}
\Require Estimated bounding box $B$, vehicle size $\langle width, height \rangle$, roadside LiDAR location $M$
\Ensure size corrected localization result $C$
\State $P\gets$ nearest vertex in $B$ to $M$\Comment{The alignment point}
\State $P_l, P_r\gets$ two adjacent vertexes of $P$ in $B$
\State $\bm{L}\gets$\Call{Normalized}{$P_l - P$}
\State $\bm{R}\gets$\Call{Normalized}{$P_r - P$}
\If{$\|P_l - P\| < \|P_r - P\|$}
    \State $C\gets P + (\bm{L}\cdot width + \bm{R}\cdot height)/ 2$
\Else
    \State $C\gets P + (\bm{R}\cdot width + \bm{L}\cdot height)/ 2$
\EndIf
\end{algorithmic}
\end{algorithm}

Knowing the ground truth dimensions of a vehicle's bounding box enables precise adjustments by aligning to one corner point. Among the four corners determined by the L-shape fitting algorithm, the corner closest to the roadside LiDAR is the most reliable. A similar idea can also be found in~\cite{Wu2018-hk}. We designate this point as the \textit{alignment point}. Aligning the ground truth bounding box with this corner allows us to refine the vehicle's position. The detailed algorithm is listed in Algorithm~\ref{alg:size_correction}.

\begin{figure}
\centering
\includegraphics[width=.95\linewidth]{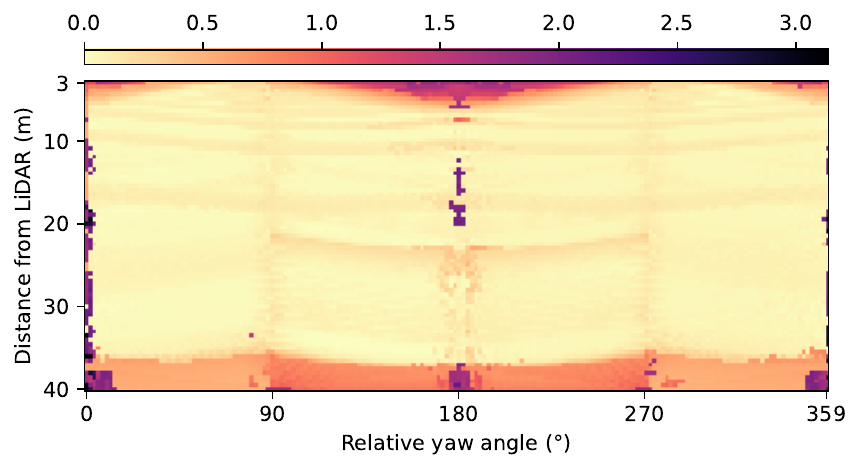}
\caption{Error of the Center Point in Distance (m)\\after bounding box size correction.}
\label{fig:error_dist_size_corrected}
\end{figure}

After applying the size correction algorithm, we ran the same pilot experiment and regenerated a new error heat map in Figure~\ref{fig:error_dist_size_corrected}. Upon comparing with Figure~\ref{fig:error_dist_original}, it becomes evident that although accuracy degraded in some instances, the overall precision markedly improved. Specifically, between the distances of 6 m and 36 m, we achieved an accuracy within 0.1 m for the majority of angles, satisfying the requirements for vehicle localization. When comparing Figure~\ref{fig:error_dist_size_corrected} with Figure~\ref{fig:error_in_angle}, the off-by-90° error happens when the places where the error increases.

\section{Realistic Experiment and Evaluation}\label{experiment}
\subsection{Experiment Setup}

\begin{figure}[h]
    \centering
    \includegraphics[width=.9\linewidth]{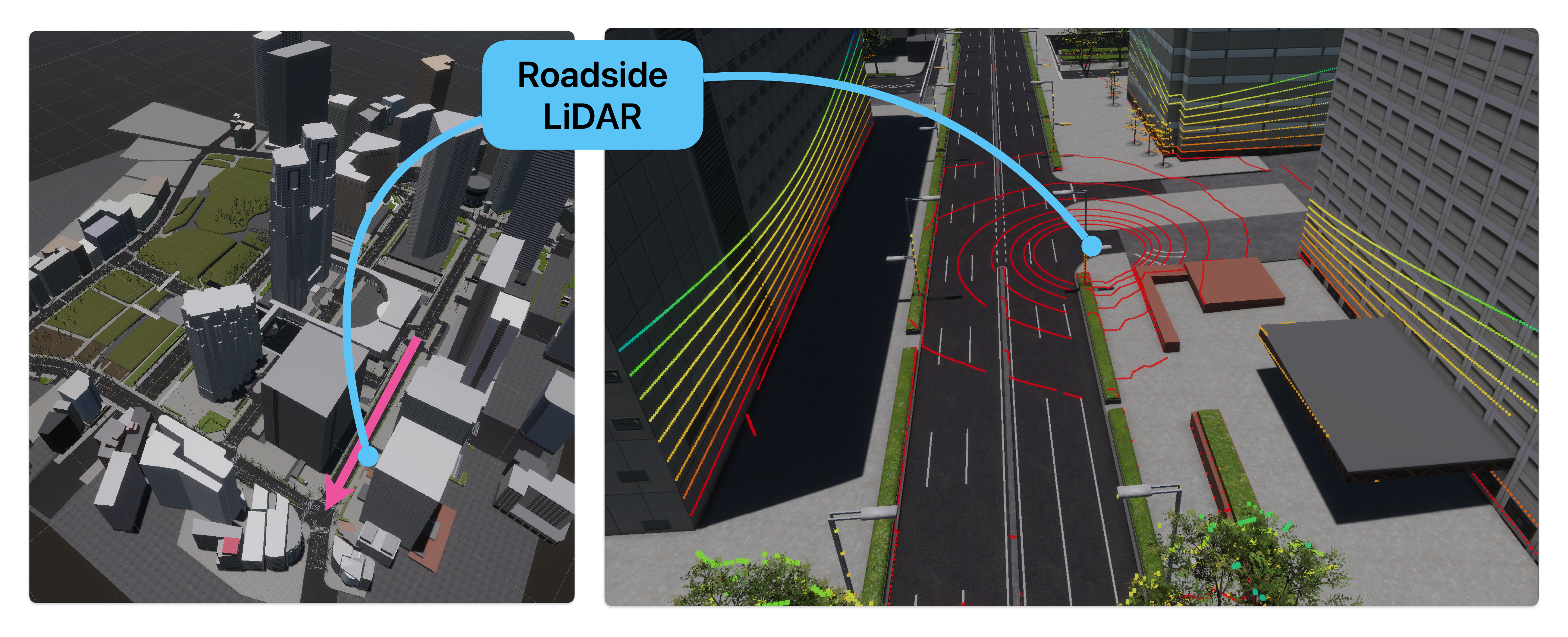}
    \caption{The bird view of the Nishishinjuku Area in AWSIM. The pink arrow is where we conducted the realistic experiment, and the blue dot is where we put the roadside LiDAR.}
    \label{fig:bird_view}
\end{figure}

To evaluate the efficacy of our proposed method in a practical setting, we employed an  Autoware and AWSIM co-simulation environment. For this experiment, the AWSIM map is modeled based on the Nishishinjuku district in Tokyo, Japan. Figure~\ref{fig:bird_view} offers an aerial perspective of the 3D test environment. The 3D model and the HD Point Cloud Map are provided by Tier IV\footnote{https://tier4.jp/en/}, which are required to perform NDT scan matching-based localization.

In the Autoware localization framework, an Extended Kalman Filter (EKF) is used to fuse NDT scan matching with other odometer. However, in AWSIM, the Global Navigation Satellite System (GNSS) data and the onboard Inertial Measurement Unit (IMU) consistently report accurate results, fusing these additional data sources would result in unrealistically accurate localization. Thus we turned off input sources other than NDT scan matching for a realistic evaluation. Both the vehicle-mounted and roadside LiDAR systems operate at a frequency of 10 Hz, thus delivering localization outcomes at the same rate.

First, we drove the vehicle across the Nishishinjuku map without using roadside LiDAR to assess self-localization performance in various areas. We observed a decline in localization accuracy on the specific road shown in Figure~\ref{fig:bird_view}. We hypothesize that this reduction in accuracy could be due to the road being flanked by two tall buildings, limiting the vehicle's ability to capture longitudinal features and thereby introducing localization error~\cite{Javanmardi2018-us}. So, we decided to deploy our roadside LiDAR in the middle of the aforementioned road's sidewalk. The LiDAR was set up at a height of 2 m and oriented with zero roll and pitch angles.

\begin{figure}[h]
    \centering
    \includegraphics[width=.8\linewidth]{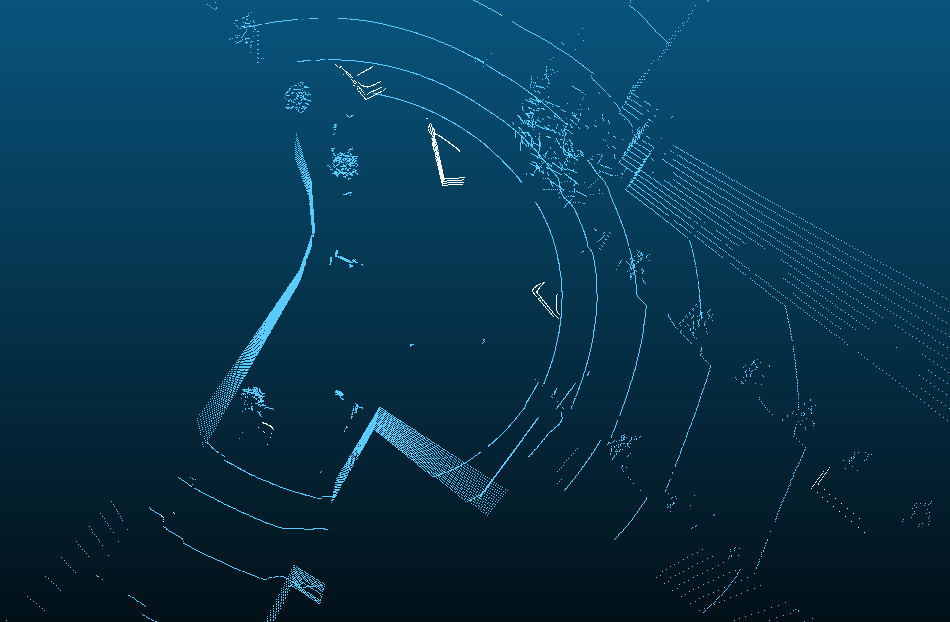}
    \caption{Background filtering. The blue points are background points; the white points are vehicle points.}
    \label{fig:background_filtering}
\end{figure}

For background filtering, we first record a reference frame from the LiDAR while ensuring that the vehicle is outside its visible range. When the LiDAR subsequently senses a vehicle, it will compare the current LiDAR frame with the reference frame. Points appearing in the current frame but absent in the reference frame are attributed to the vehicle. We used K-D tree to find corresponding points between the current and reference frames, which enables us to segregate the vehicle points from the background points. Figure~\ref{fig:background_filtering} shows an example of the background filtering process.

\begin{figure}[h]
    \centering
    \includegraphics[width=\linewidth]{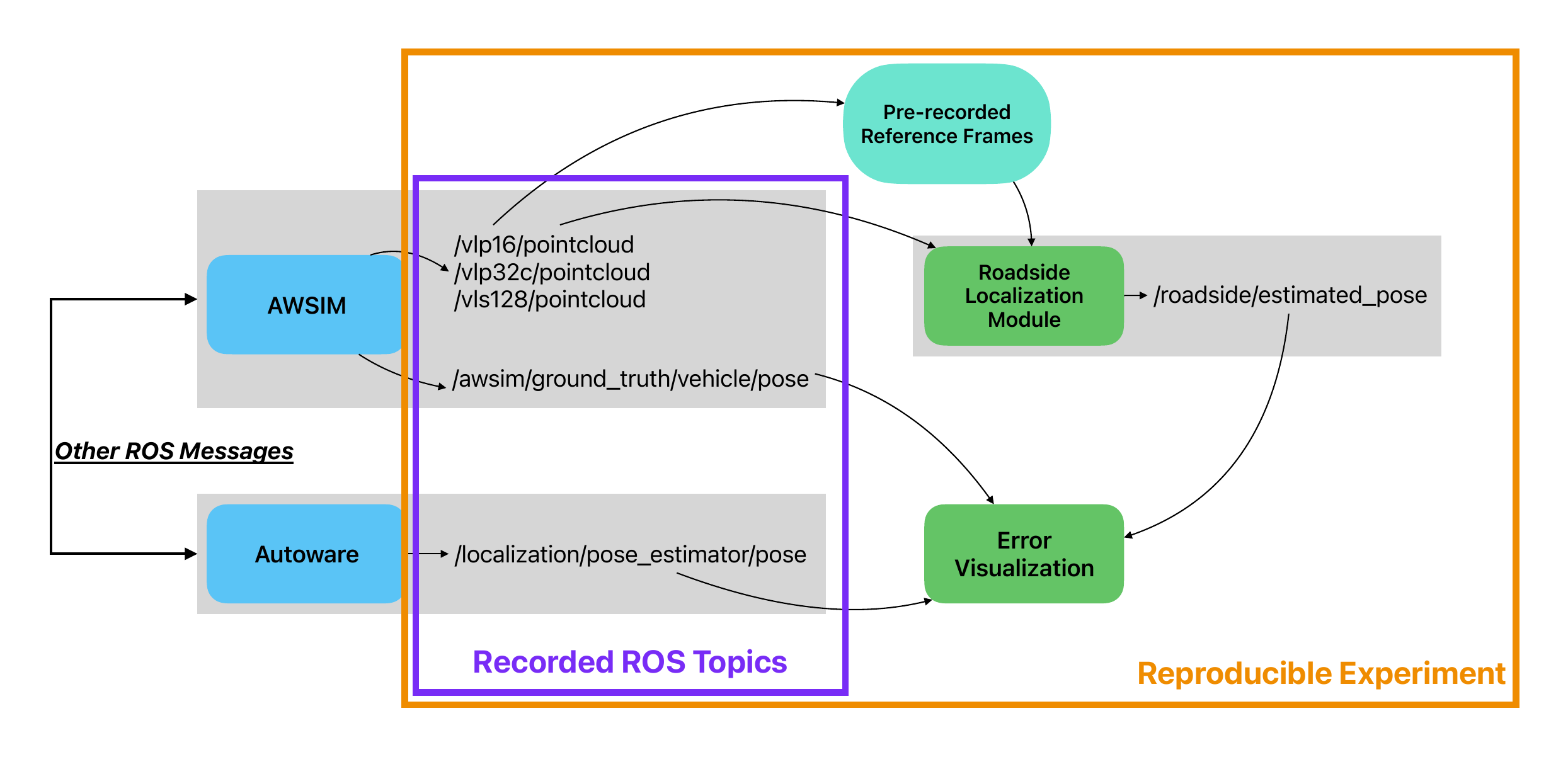}
    \caption{The data flow of the experiment}
    \label{fig:dataflow}
\end{figure}

As illustrated in Figure~\ref{fig:dataflow}, our experimental setup is organized in the following manner. We initially position multiple overlapping LiDAR models at the same location and record their individual reference frames. Subsequently, we activate AWSIM-Autoware cosimulation and manually navigate the vehicle along the designated road.  Meanwhile, we capture the \texttt{rosbag}, containing the roadside LiDARs' raw data, the ground truth of the vehicle's position, and Autoware's localization results. Because the vehicle is manually controlled, we cannot guarantee the exact input for each trial if we want to execute the same experiment multiple times; replaying the \texttt{rosbag} allows us to reproduce the same conditions across multiple trials. We establish a roadside localization module programmed to listen to raw point cloud output topics, with pre-existing knowledge of the reference frames and the vehicle's bounding box dimensions. Upon replaying the recorded \texttt{rosbag}, we aggregate position data from the ground truth, vehicle self-localization, and roadside localization modules for visualization.

\begin{figure}
    \centering
    \includegraphics[width=0.9\linewidth]{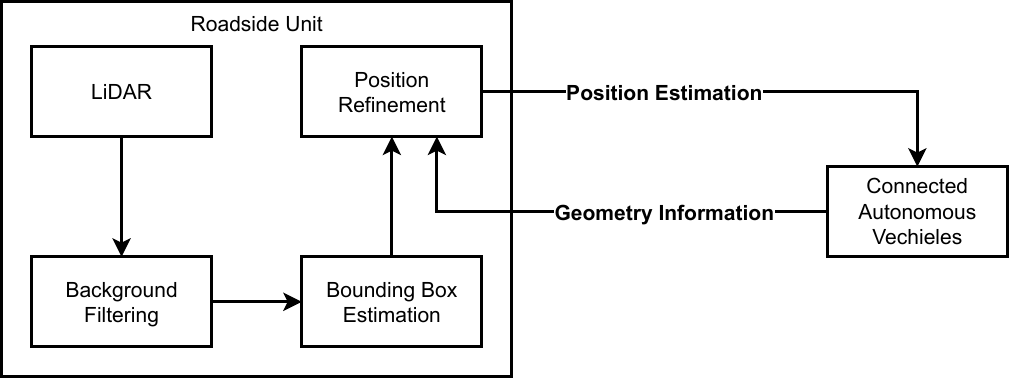}
    \caption{Flowchart of proposed method}
    \label{fig:flowchart}
\end{figure}

\begin{figure}[h]
    \centering
    \includegraphics[width=\linewidth]{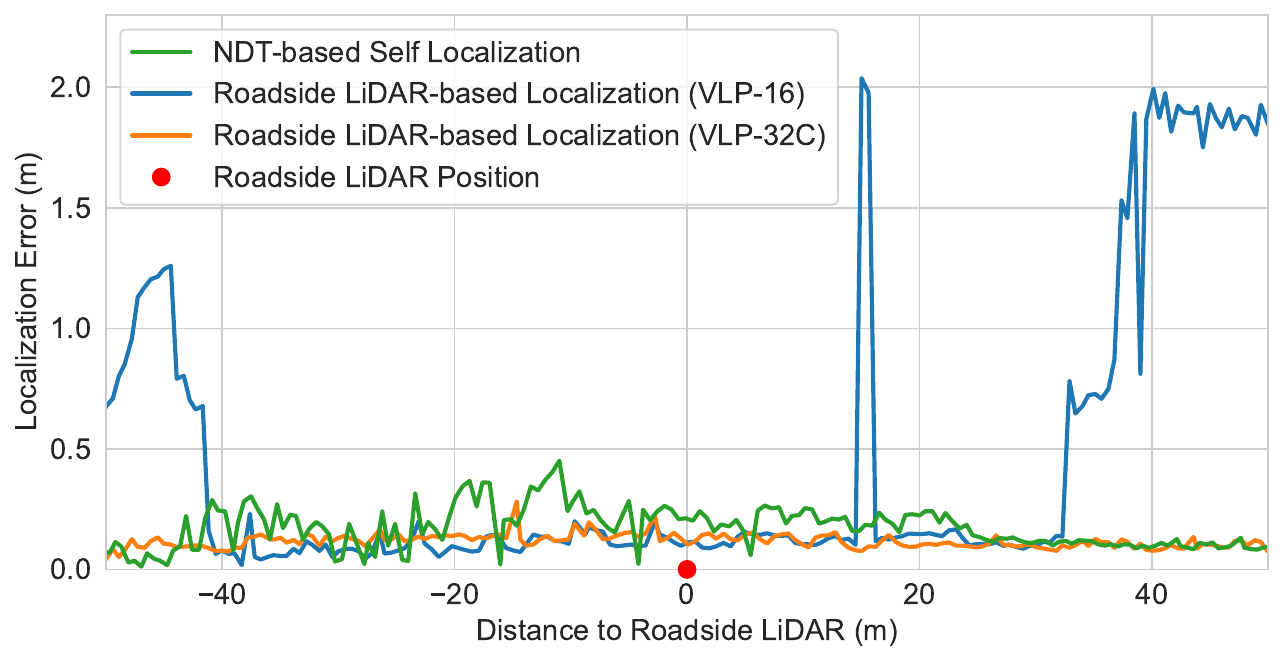}
    \caption{Localization error comparison}
    \label{fig:comparison}
\end{figure}

\subsection{Analysis and Evaluation}\label{analysis}

In the experimental results Figure~\ref{fig:comparison}, we can clearly see that our proposed method has significant improvement over the NDT-based localization on the target road in the effective range. The two outlier points at around 17m in the graph were due to the off-by-90° issue, as discussed in Section~\ref{init}. This did not happen in the pilot experiment because there is no occlusion in the pilot experiment. However, in a real-world setting, various occlusions like shrubbery, street lamps, and elevated roadways can hinder the LiDAR's ability to capture points, thereby making our method more susceptible to errors. However, if we use a more powerful LiDAR model, VLP-32C, no extreme error will happen on the target road, enhancing both stability and accuracy.

\begin{figure}[h]
    \centering
    \includegraphics[width=\linewidth]{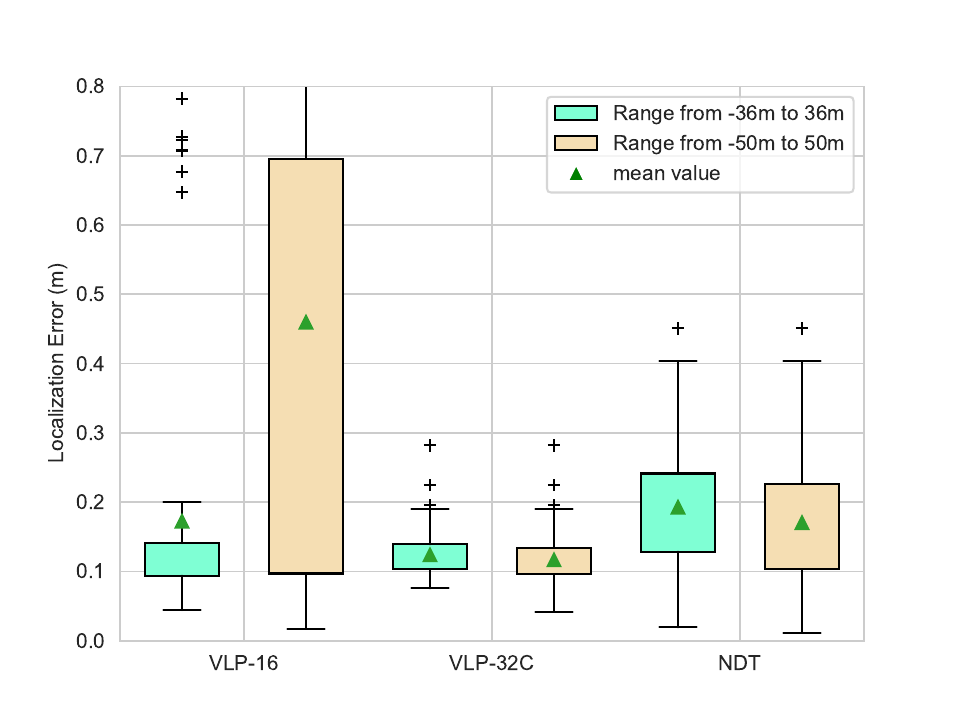}
    \caption{Box plot of error from VLP-16, VLP-32C, and NDT-based localization. The boxes represent 25th percentile to 75th percentile. Note that since the figure is cropped, the data for VLP-16 is not completely shown.}
    \label{fig:box}
\end{figure}

We further examine the data in the effective range for the two LiDAR models presented in Figure~\ref{fig:box}. In the range from -36 m to 36 m, both the VLP-16 and VLP-32C models consistently exhibit sub-30 cm level accuracy. Notably, the VLP-32C model extends this high level of accuracy across a broader range, from -50 m to 50 m, outperforming NDT-based localization in a statistical sense.

\begin{table}[h]
\centering
\caption{The mean absolute error of NDT-based localization and the proposed method with 2 different LiDAR models.}
\begin{tabular}{@{}cccc@{}}
\toprule
\multirow{2}{*}{Range} & \multirow{2}{*}{\begin{tabular}[c]{@{}c@{}}NDT-based\\ Self Localization\end{tabular}} & \multicolumn{2}{c}{Proposed Method} \\ \cmidrule(l){3-4} 
                       &                                                                                        & VLP-16           & VLP-32C          \\ \midrule
-36 m to 36 m            & 0.1927                                                                                 & 0.1724           & 0.124            \\
-50 m to 50 m            & 0.1704                                                                                 & 0.4596           & 0.1167           \\ \bottomrule
\end{tabular}
\end{table}

\section{Conclusion}\label{conclusion}
In this paper, we proposed a roadside LiDAR-based cooperative localization method for connected vehicles, aimed at enhancing localization accuracy where onboard vehicle self-localization falls short. An initial pilot experiment was conducted to gauge the effectiveness of roadside LiDAR in vehicle perception and identify possible sources of errors. We then introduced a novel method to let connected vehicles send their dimensions to the RSU via V2I communication. This enables the RSU to provide a more accurate localization result for these connected vehicles. The experimental result of the realistic simulation confirms that our proposed method yields better localization results than NDT scan matching-based vehicle self-localization within the effective range.


This research is in its nascent phase, offering ample opportunities for further exploration. One limitation is the focus on single-vehicle scenarios; future work should address the added complexities when multiple vehicles are present, including increased occlusion.

Besides, as outlined in Section~\ref{analysis}, the off-by-90° issue results in peaks of localization error. Although it can be mitigated by using more powerful LiDAR models, it does not entirely eliminate the possibility of its occurrence. More effort should be made to mitigate this problem to improve the overall stability.

Lastly, the LiDAR position significantly impacts the localization result in our proposed method. Owing to paper length constraints, we have only examined a limited number of positioning strategies for the roadside LiDAR. Other researchers have also recently drawn attention to the importance of roadside LiDAR positioning~\cite{Cai2023-ln}. We also want to explore how the position of roadside LiDAR will affect our proposed method in the future.

\section*{Acknowledgments}
This work was supported by the Japan Society for the Promotion of Science (JSPS) KAKENHI (grant number: \#21H03423).

\bibliographystyle{ACM-Reference-Format}
\bibliography{main}
\end{document}